\documentclass{tlp}
\usepackage{amssymb}
\usepackage{graphicx}
\usepackage{hyperref} 
\usepackage{capt-of}
\usepackage{url}

\makeatletter
\newif\if@restonecol
\makeatother

\usepackage[ruled,figure]{algorithm2e}
\usepackage{amsmath}
\usepackage{amsfonts}
\usepackage{listings}
\usepackage{color}

\newcommand{\comment}[1]{}
\newcommand{\oursystem}{\emph{{CDF-Rules}}}
\newcommand\MKNF{{\em MKNF$_{WFS}$}}

\newcommand{\cI}{{\cal I}}
\newcommand{\cK}{{\cal K}}
\newcommand{\cO}{{\cal O}}
\newcommand{\longline}{\noindent\rule{\textwidth}{.01in}}

\newtheorem{definition}{Definition}

\newtheorem{example}{Example}

 \submitted{1 June 2010}
 \revised{18 October 2010}
 \accepted{25 July 2012}

\begin{document}


\title[A Goal-Directed Implementation of Query Answering for Hybrid MKNF]{A Goal-Directed Implementation of Query Answering for Hybrid MKNF Knowledge Bases}

\author[A S Gomes, J J Alferes and T Swift]{Ana Sofia Gomes, Jos\'e J\'ulio Alferes  and Terrance Swift\\
      CENTRIA, Departamento de Inform\'atica\\
       Faculdade Ci\^{e}ncias e Tecnologias \\
       Universidade Nova de Lisboa\\
       2829-516 Caparica, Portugal}

\maketitle

\begin{abstract}

Ontologies and rules are usually loosely coupled in knowledge
representation formalisms.  In fact, ontologies use open-world
reasoning while the leading semantics for rules use non-monotonic,
closed-world reasoning.  One exception is the tightly-coupled
framework of Minimal Knowledge and Negation as Failure (MKNF), which
allows statements about individuals to be jointly derived via
entailment from an ontology and inferences from rules.  Nonetheless,
the practical usefulness of MKNF has not always been clear, although
recent work has formalized a general resolution-based method for
querying MKNF when rules are taken to have the well-founded semantics,
and the ontology is modeled by a general oracle.  That work leaves
open what algorithms should be used to relate the entailments of the
ontology and the inferences of rules.  In this paper we provide such
algorithms, and describe the implementation of a query-driven system,
\oursystem{}, for hybrid knowledge bases combining both (non-monotonic)
rules under the well-founded semantics and a (monotonic) ontology,
represented by a CDF Type-1 ($\mathcal{ALCQ}$) theory.\\
\end{abstract}


\section{Introduction}

Ontologies and rules offer distinctive strengths for the
representation and transmission of knowledge over the Semantic Web.
Ontologies offer the deductive advantages of first-order logics with
an open domain, while guaranteeing decidability.  Rules employ
non-monotonic (closed-world) reasoning that can formalize scenarios
under locally incomplete knowledge; rules also offer the ability to
reason about fixed points (e.g. reachability)  which cannot be
expressed within first-order logic.

\begin{example}\label{motivEx}
Consider a scenario application for a Customs Agency where an ontology
is used to assess and classify aspects of imports and exports. An
ontology with such characteristics would embrace several thousand
axioms. In these axioms, as an example, one can define several axioms
about countries (e.g. Scandinavian countries are European countries;
Norway is a Scandinavian country; etc). Moreover, one can also model
in the ontology knowledge that is certain, in the sense that it does
not allow for exception. With such
certain knowledge, first-order logics deduction is what is
desired. For example, in this scenario, one could state that
Scandinavian countries are always considered safe countries. All of
this can be easily represented by an ontology defined in a Description
Logic (DL) language \cite{dl-book}\footnote{We write DL formulas with
  the usual notation and the usual DL operators (see \cite{dl-book}):
  so that the argument variable of a unary predicate is not displayed
  and the first letter of the predicate's name is capitalized.}
(\ref{safecountryMot}).
\begin{equation}\label{safecountryMot}
\begin{array}{l}
ScandinavianCountry \sqsubseteq EuropeanCountry\\ Norway:ScandinavianCountry \\
ScandinavianCountry \sqsubseteq SafeCountry\\
\end{array}
\end{equation}

Besides these axioms about countries, one may want to specify some
additional knowledge, e.g. for defining conditions about whether or
not to inspect entering shipments based on their country of
origin. For example, we may want to state that one should inspect any
vessel containing 
a shipment coming from a country that is not guaranteed to
be safe. Note here that 
the
closed world assumption is needed to define
such a statement. In fact, intuitively one wants 
to assume that, if one is unsure whether given country is safe, then
(at least for the sake of this statement) the country should be
considered unsafe and an inspection performed.  When using description
logic with classical negation this behavior would not be obtained, as
a condition $\neg SafeCountry(country)$ would only be true in case one
knows for sure that the country is not safe. However, this statement
can be easily expressed by using a (non-monotonic) rule with default
negation\footnote{We write rules in the usual notation for logic
  programming, where predicates names are non-capitalized (even if
  common to the ontology, where they are capitalized), and variables
  are capitalized and implicitly universally quantified in the rule head.}
as in (\ref{inspectMot1}).
%
\begin{equation}\label{inspectMot1}
inspect(X) \leftarrow hasShipment(X,Country), \mathbf{not} \ safeCountry(Country).
\end{equation}

Such rules are non-monotonic in the sense that further knowledge -- in
this case about safety of countries -- can invalidate previous
conclusions about inspection. This kind of non-monotonic rule is quite
useful to specify default knowledge that may be subject to
exceptions. In this sense rule (\ref{inspectMot1}) can be seen as
stating that by default shipments should be inspected, but an
exception to this default rule are shipments coming from safe
countries.

Note that defining some statements about a predicate using default
rules does not necessarily mean that {\em all\ } statements defining
that predicate should be default rules. For example, we may want to state
that diplomatic shipments are not subject to inspection, regardless
their country of origin. Here no default reasoning is
involved. Moreover, negation here is not the default negation of logic
programming, but rather classical negation.
Such a statement could be modeled using normal logic programs extended with explicit negation \cite{answer-sets} with the rule:
\begin{equation}\label{explicitnegrule}
\begin{array}{l}
\neg inspect(X) \leftarrow diplomaticShipment(X)
\end{array}
\end{equation}
\noindent or, having the possibility to use an ontology
formalized as a fragment of first-order logic, with a statement:
\begin{equation}\label{govships}
\begin{array}{l}
DiplomaticShipment \sqsubseteq \neg Inspect
\end{array}
\end{equation}

However, these two representations of the statement above are not
equivalent. In particular, with axiom \ref{govships} one can conclude
that a given shipment is not diplomatic whenever one knows, by some
other rule, that it should be inspected; this is not the case with
rule \ref{explicitnegrule}.  The behavior of rules with explicit
negation can be easily modeled by using ontologies; for that it
suffices to create a new concept standing for the explicit negation,
say $NonInspect$, replacing the explicitly negated literal in the rule
by this new concept, and adding the axiom $NonInspect \equiv
\neg Inspect$. On the contrary, with programs extended with explicit
negation one cannot obtain the full expressivity that is obtained with some of the DL-based languages for
ontologies, simply because some of these languages belong to a higher complexity class.
Another important feature not supported by explicit
negation but typically supported by DL-based ontologies, is the
possibility of using existential quantification as illustrated in
Example \ref{ex:discount} below.
\end{example} 

The foregoing example briefly illustrates some advantages of combining features of ontologies with features of logic programming-like rules, and of doing so in a way that allows knowledge about instances to be
fully inter-definable between rules and an ontology (as happens with the predicate $inspect/1$ above, which is both defined in the rules and the ontology).

In fact, this combination of ontology and rule languages has gained particular importance in the context of the Semantic Web \cite{HMRS06}. In this context, a family of languages for representing background knowledge in the Web, OWL 2 \cite{OWL_Primer},  has been recommended by the World Wide Web Consortium (W3C). OWL 2 languages are based on Description Logics \cite{dl-book}, which, as in the example above, employ open-world assumption. 
In addition, the rule interchange language RIF \cite{RIF_Primer} has
recently been formally recommended to the W3C.  RIF has many
similarities with the logic programming language used in the example
above, and adopts the closed-world assumption.

The existence of both rules and ontology languages should make it possible to combine open and closed world reasoning, and this combination is indeed important in several domains related to the Semantic Web.
As a further example where this combination is desired, consider the large case study described in \cite{PatelEtAl07}, containing millions of assertions concerning matching patient records to clinical trials criteria.
In this case study, open world reasoning is needed in deductions about domains such as radiology and laboratory data: unless a lab or radiology test asserts a negative finding no arbitrary assumptions about the results of the test can be made (e.g. we can only be certain that some patient does not have a specific kind of cancer if the corresponding test is known to have negative result).
However, as observed in  \cite{PatelEtAl07}, closed world reasoning can and should be used with pharmacy data to infer that a patient is not on a medication if this is not asserted. 
The work of \cite{PatelEtAl07} applies only open world reasoning but claims that the usage of closed world reasoning in pharmacy data would be highly desirable.
Similar situations occur e.g. in matchmaking using Semantic Web Services (cf. \cite{GH:SMatch}), where again a combination of ontology languages relying on open-world reasoning, with rule languages relying on closed-world reasoning is considered highly desirable.

Several factors influence the decision of how to combine rules and
ontologies into a hybrid knowledge base.  The choice of semantics for
the rules, such as the answer-set semantics~\cite{answer-sets} or the
well-founded semantics (WFS)~\cite{GRS91}, can greatly affect the
behavior of the knowledge base system.  The answer-set semantics
offers several advantages: for instance, description logics can be
translated into the answer-set semantics providing a solid basis for
combining the two paradigms~\cite{Bara02,EiterLST04,Swif04,Mot06}.  On
the other hand, WFS is weaker than the answer-set semantics (in the
sense that it is more skeptical), but has the advantages of lower
complexity and its ability to be evaluated in a query-oriented
fashion, which have led to its integration into Prolog
systems. Another possibility of maintaining the complexity under
reasonable bounds for Semantic Web applications, is to limit the
expressivity of both the ontological part of the knowledge base, and
of the rules. For instance, \cite{DBLP:journals/ws/CaliGL12} considers
variants of Datalog that allow for existential quantification in rule
heads along with other features to support a restricted form of
ontological reasoning, yet restrict the rule syntax to obtain
tractability.

Keeping to the general form of logic programming rules, but
maintaining the complexity of reasoning under reasonable bounds,
several formalisms have concerned themselves with combining ontologies
with general WFS rules~\cite{DM07,ELST04:RuleML,KAH:ECAI08}.
Among these, the Well-Founded Semantics for Hybrid MKNF knowledge
bases (\MKNF{}), introduced in \cite{KAH:ECAI08} and overviewed in
Section~\ref{s:prel} below, is the only one that allows knowledge
about instances to be fully inter-definable between rules and an
ontology that is taken as a parameter of the formalism.  \MKNF{}
assigns a well-founded semantics to Hybrid MKNF knowledge bases, is
sound w.r.t. the original semantics of \cite{MotikR07} and, as
in~\cite{MotikR07}, allows the knowledge base to have both closed- and
open-world (classical) negation.  For comparisons of \MKNF{} with the
approaches for combining rules with ontologies mentioned above, see
\cite{DBLP:journals/ai/KnorrAH11}. 

\begin{example} \rm \label{ex:discount}
The following fragment, adapted from an example in \cite{MotikR07},  concerning car insurance
premiums illustrates several properties of
\MKNF .  The ontology consists of the following axioms, 
which state that $Married$ and $NonMarried$ are complementary concepts, that anyone who is not married is high risk, and that anyone with a spouse is married:
\begin{align*}
NonMarried \equiv \neg Married \\
\neg Married \sqsubseteq HighRisk \\ 
\exists Spouse.\top \sqsubseteq Married.
\end{align*}
M\noindent 
The rule base consists of the following rules, which state that anyone who is not known to be married is to be assumed to be non-married, that those who are known to be high-risk should have a surcharge, and that those that have a known spouse should have a discount:

\begin{align*}
&\mathbf{K}\ nonMarried(X) \leftarrow \mathbf{K}\ person(X), \mathbf{not}\ married(X). \\
&\mathbf{K}\ surcharge(X) \leftarrow \mathbf{K}\ highRisk(X),  \mathbf{K}\ person(X).\\
&\mathbf{K}\ discount(X) \leftarrow \mathbf{K}\ spouse(X,Y),  \mathbf{K}\ person(X), \mathbf{K}\ person(Y). 
\end{align*}

Note that {\em married} and {\em nonMarried} are defined both by
axioms in the ontology and by rules.  Within the rule bodies, literals
with the {\bf K} or {\bf not} operators
(e.g. $\mathbf{K}\ highRisk(X)$) may require information both from the
ontology and from other rules; other literals are proven directly by
the other rules (e.g. {\em person(X)}). Intuitively, $\mathbf{K} \varphi$ stands for ``$\varphi$ is known to be true'', whereas $ \mathbf{not}\ \varphi$ stands for ``$\varphi$ is not known to be true".

Suppose {\em person(john)} were added as a fact (in the rule base).
Under closed-world negation, the first rule would derive {\em
  nonMarried(john)}.  By the first ontology axiom, 
{\em $\neg$married(john)} would hold, and by the second axiom {\em
  highRisk(john)} would also hold.  By the last rule, $
surcharge(john)$ would hold as well.  Thus the proof of $surcharge(john)$
involves interdependencies between the rules with closed-world
negation, and the ontology with open-world negation.  At the same time
the proof of $surcharge(john)$ is {\em relevant} in the sense that
properties of other individuals, not related to $john$ either through rules nor axioms in the ontology, do not need to be considered.


Now suppose that one learns that a person named Bill has a spouse, and
that Bob is the spouse of Ann. This can be formalized by adding the
corresponding facts for the predicate $person$, the fact
$spouse(bob,ann)$ and the DL assertion $bill:\exists Spouse.\top$. In
this case one would expect that neither Bill nor Bob are considered
high risk, and so should not have a surcharge; and that since Bob
(contrary to Bill) has a known spouse, he should have a discount. This
is in fact the result of \MKNF. Note that, it is only possible to
represent the difference between the situations of Bill and Bob by
using the existential quantification of DL, something that is not
possible in logic programming alone.
\end{example}

In the original definition of \MKNF{}, the inter-dependencies of the
ontology and rules were captured by a bottom-up fixed-point operator
with multiple levels of iterations.  Recently, a query-based approach
to hybrid MKNF knowledge bases, called SLG($\cO$), has been developed
using tabled resolution~\cite{AlKS12}.  SLG($\cO$) is sound and complete, as well as
terminating for various classes of programs
(e.g. datalog).  In addition SLG($\cO$) is relevant in
the sense of Example~\ref{ex:discount}, i.e. in general one does not need to compute the whole model (for every object in the knowledge base) to answer a specific query.
This relevancy is a critical requirement for scalability in numerous
practical applications (e.g. in the area of Semantic Web): without
relevance a query about a particular individual $I$ may need to derive
information about other individuals even if those individuals were not
connected to $I$ through rules or axioms. This is also clear e.g. in
the above mentioned case study about matching patient records for
clinical trials criteria \cite{PatelEtAl07}.  In this study one is
interested in finding out whether a given patient matches the criteria
for a given trial or, at most, which patients match a given criteria;
for scalability, it is crucial that such a query does not need to go
over all patients and all criteria.  In a similar manner, when
assessing a shipment in Example~\ref{motivEx}, it is infeasible to
have to consider all shipments into a country: the query must be
relevant to be useful.

SLG($\cO$) serves as a theoretical framework for query evaluation of
\MKNF{} knowledge bases, but it models the inference mechanisms of an ontology
abstractly, as an oracle.  While this abstraction allows the
resolution method to be parameterized by different ontology formalisms
in the same manner as \MKNF{}, it leaves open details of how the
ontology and rules should interact and these details must be accounted
for in an implementation.

This paper describes, in Section~\ref{sec:description}, the design and
implementation of a 
prototype query evaluator\footnote{The
  implementation is available from the XSB CVS repository as part of
  the CDF package in the subdirectory {\tt packages/altCDF/mknf}.} for
\MKNF, called \oursystem{}, which fixes the ontology part to
$\mathcal{ALCQ}$ theories, and makes use of the prover from XSB's
ontology management library, the Coherent Description Framework
(CDF)~\cite{cdf} (overviewed in Section~\ref{xsb}). To the best of our
knowledge, this implementation is the first working query-driven
implementation for Hybrid MKNF knowledge bases, combining rules and
ontology and complete w.r.t. the well-founded semantics.

\section{MKNF Well-Founded Semantics}\label{s:prel}

\newcommand{\K}{\mathord\mathbf{K\ }}
\newcommand{\nf}{\mathord\mathbf{not\ }}
\newcommand{\lfp}{\operatorname{\sf lfp}}

Hybrid MKNF knowledge bases, as introduced in \cite{MotikR07}, are
essentially formulas in the logics of minimal knowledge and negation
as failure (MKNF) \cite{Lif91}, i.e. first-order logics with equality
represented as the congruence relation $\approx $, and two modal
operators $\K$ and $\nf$, which allow inspection of the knowledge
base.  Intuitively, given a first-order formula $\varphi$, $\K
\varphi$ asks whether $\varphi$ is known, while $\nf \varphi$ 
checks whether $\varphi$ is not known.  A Hybrid MKNF knowledge base
consists of two components: a decidable description logic (DL)
knowledge base, translatable into first-order logic; and a finite set
of rules of modal atoms.

\begin{definition}\label{d:MKNFKB}
Let $\mathcal{O}$ be a DL knowledge base built over a language $\mathcal{L}$ with distinguished sets of countably infinitely many variables $N_{V}$, along with finitely many individuals $N_{I}$ and predicates (also called concepts) $N_{C}$.
An atom $P(t_{1},\ldots,t_{n})$, where $P\in N_{C}$ and $t_{i}\in N_{V}\cup N_{I}$, is called a \emph{DL-atom} if $P$ occurs in $\mathcal{O}$,  otherwise it is called \emph{non-DL-atom}.

An MKNF rule $r$ is of the form
\begin{equation}\label{mknfRule}
\K H\leftarrow \K A_{1},\ldots, \K A_{n}, \nf B_{1},\ldots,\nf B_{m}
\end{equation}
where $H_{i}$, $A_{i}$, and $B_{i}$ are atoms.
$H$ is called the \emph{(rule) head} and the sets $\{\K A_{i}\}$, and $\{\nf B_{j}\}$ form the 
\emph{(rule) body}.
Atoms of the form $\K A$ are also called \emph{positive literals} or \emph{modal $\mathbf{K}$-atoms} while atoms of the form $\nf A$ are called \emph{negative literals} or \emph{modal $\mathbf{not}$-atoms}. 
A rule $r$ is \emph{positive} if $m=0$ and a \emph{fact} if $n=m=0$. 
 A \emph{program} $\mathcal{P}$ is a finite set of MKNF rules and 
 a \emph{hybrid MKNF knowledge base} $\mathcal{K}$ is a pair $(\mathcal{O},\mathcal{P})$.
\end{definition}

In Definition~\ref{d:MKNFKB}, the modal operators of MKNF logics are
only applied to atoms in the rules, which might indicate that it would
suffice to use a simpler logic that does not deal with the application
of the modal operators to complex formulae. However, for defining the
meaning of these hybrid knowledge bases, modal operators do need to be
applied to complex formulas, namely to the whole ontology
$\mathcal{O}$ (cf. Definition \ref{d:MKNFWFS}). Intuitively, a hybrid
knowledge base specifies that: for each rule of the form
(\ref{mknfRule}) in $\mathcal{P}$, the atom $H$ is known if all the
$A_i$ are known and none of the $B_j$ is known; and that the whole
ontology $\mathcal{O}$ is known (i.e. $\K \mathcal{O}$). As such,
using the more general MKNF logic eases the definition of the
semantics of hybrid knowledge bases,  so similarly to \cite{MotikR07}
and \cite{KAH:ECAI08}, we resort to the full MKNF logic \cite{Lif91}.

\comment{
Original:

In this definition of hybrid knowledge bases, the modal operators of
the logics MKNF are only applied to atoms in the rules, which might
indicate that it would suffice to use a logic that does not deal with
the application of the modal operators to complex formulae. However,
for defining the meaning of these knowledge bases, modal operators are
also applied to more complex formulas, namely to the whole ontology
$\mathcal{O}$ (cf. Definition \ref{d:MKNFWFS}). Intuitively, a hybrid
knowledge base specifies that: atom $H$ is known if all the $A_i$ are
known and none of the $B_j$ is known for each rule of the form
(\ref{mknfRule}) in $\mathcal{P}$; and that the whole ontology
$\mathcal{O}$ is known (i.e. $\K \mathcal{O}$). As such, using a more
general logic, like MKNF, eases the definition of the semantics of
hybrid knowledge bases and, similarly to what is done in
\cite{MotikR07} and \cite{KAH:ECAI08}, we resort to the full logics
MKNF \cite{Lif91} for that definition.
}

For decidability DL-safety is assumed,
which basically constrains the use of rules to individuals actually appearing in the knowledge base under consideration.
Formally, an MKNF rule $r$ is \emph{DL-safe} if every variable in $r$ occurs in at least one non-DL-atom $\K B$ occurring in the body of $r$.
A hybrid MKNF knowledge base $\mathcal{K}$ is \emph{DL-safe} if all its rules are DL-safe (for more details we refer to \cite{MotikR07}).

The well-founded MKNF semantics, \MKNF{}, 
presented in
\cite{KAH:ECAI08}, and further developed in \cite{DBLP:journals/ai/KnorrAH11}, is based on a complete three-valued extension of
the original MKNF semantics.  However here, as we are only interested
in querying for literals and conjunctions of literals, we limit
ourselves to the computation of what is called the well-founded
partition in \cite{KAH:ECAI08}: basically the atoms that are true or
false.  For that reason, and in correspondence to logic programming,
we name this partition the well-founded model.  
First, we recall
some notions from \cite{KAH:ECAI08} which will be useful in the
definition of the operators for obtaining that well-founded model.

$ $\\

\begin{definition}\label{d:katoms}
Consider a hybrid MKNF knowledge base $\mathcal{K}=(\mathcal{O},\mathcal{P})$.
The \emph{set of $\mathbf{K}$-atoms of $\mathcal{K}$}, written $\sf{KA}(\mathcal{K})$, is the smallest set that contains (i) all modal $\mathbf{K}$-atoms occurring in $\mathcal{P}$, and (ii) a modal atom $\K \xi$ for each modal atom $\nf \xi$ occurring in $\mathcal{K}$.

Furthermore, for a set of modal atoms $S$, $S_{DL}$ is the subset of
DL-atoms of $S$ (Definition~\ref{d:MKNFKB}), and
$\widehat{S}=\{\xi\mid \K \xi \in S\}$.
\end{definition}
Basically $\sf{KA}(\mathcal{K})$ collects all modal atoms of
predicates appearing in the rules, and $\widehat{S}$ just removes
$\mathbf{K}$ operators from the argument set $S$.

To guarantee that all atoms that are false in the ontology are also
false by default in the rules, we introduce new positive DL atoms
that represent first-order false DL atoms, along with a program
transformation making these new modal atoms available for reasoning
with the respective rules.

\begin{definition}\label{d:kb+*}
Let $\mathcal{K}$ be a DL-safe hybrid MKNF knowledge base.
We obtain $\mathcal{K}^{+}$ from $\mathcal{K}$ by adding an axiom $\neg P \sqsubseteq \textit{NP}$ for every DL atom $P$ which occurs as head in at least one rule in $\mathcal{K}$ where $\textit{NP}$ is a new predicate not already occurring in $\mathcal{K}$.
Moreover, we obtain $\mathcal{K}^{*}$ from $\mathcal{K}^{+}$ by adding $\nf\ NP(t_{1},\ldots,t_{n})$ to the body of each rule with a DL atom $P(t_{1},\ldots,t_{n})$ in the head.
\end{definition}

In $\mathcal{K}^{+}$, $\textit{NP}$ represents $\neg P$ (with its
corresponding arguments) while $\mathcal{K}^{*}$ introduces a
restriction on each rule with such a DL atom in the head, saying
intuitively that the rule can only be used to conclude the head if the
negation of its head cannot be proved\footnote{Note that
  $\mathcal{K}^{+}$ and $\mathcal{K}^{*}$ are still hybrid MKNF
  knowledge bases, so we only refer to $\mathcal{K}^{+}$ and
  $\mathcal{K}^{*}$ explicitly when it is necessary.}.
For example, 
to guarantee in Example~\ref{motivEx} that proving falsity of
$Inspect$ for some shipment (via the ontology) enforces default
negation of $Inspect$ for that shipment, one would build
$\mathcal{K}^{+}$ by adding to the ontology the axiom $\neg Inspect
\sqsubseteq N\_Inspect$ (where $N\_Inspect$ is a new symbol not
appearing elsewhere), and in $\mathcal{K}^{*}$ all original rules with
$inspect(X)$ in the head would be transformed by adding
$\nf n\_inspec(X)$ to the body.  
In this case, the rule in
(\ref{inspectMot1}) would be transformed to:

\begin{tabbing}
foooooooooooooo\=ooooooooooooo\=\kill
$\mathbf{K}\ inspect(X) \leftarrow \mathbf{K}\ hasShipment(X,Country), \mathbf{not} \ safeCountry(Country), $ \\
\> $ \mathbf{not} \ n\_inspect(X).$
\end{tabbing}

We continue by recalling the definition in  \cite{KAH:ECAI08} of  an operator $T_{\mathcal{K}}$ to allow conclusions to be drawn from positive hybrid MKNF knowledge bases (i.e. knowledge bases where rules have no default negation).

\begin{definition}\label{d:opRkDkTk}
For $\mathcal{K}$ a positive DL-safe hybrid MKNF knowledge base, $R_{\mathcal{K}}$, $D_{\mathcal{K}}$, and $T_{\mathcal{K}}$ are defined on subsets of $\sf{KA}(\mathcal{K}^{*})$ as follows:
\[\begin{array}{rcl}
R_{\mathcal{K}}(S) &=& S \cup \{\K H \mid \mathcal{K}$ contains a rule of the form (1) such that $\K A_{i}\in S\\ &&\ \ \ \ \ \  $ for each $1\leq i\leq n\}\\
D_{\mathcal{K}}(S) &=& \{\K \xi \mid \K\xi\in \sf{KA}(\mathcal{K}^{*})$ and $\mathcal{O}\cup \widehat{S}_{DL} \models \xi \} \cup\{\K Q(b_{1},\ldots,b_{n})\mid \\ && \ \ \K Q(a_{1},\ldots,a_{n})\in S\setminus S_{DL}$, $\K Q(b_{1},\ldots,b_{n})\in \sf{KA}(\mathcal{K}^{*})$, and $\\ &&\ \ \mathcal{O}\cup \widehat{S}_{DL}\models a_{i}\approx b_{i} \ $ for $1\leq i\leq n\}\\
T_{\mathcal{K}}(S) &=& R_{\mathcal{K}}(S)\cup D_{\mathcal{K}}(S)
\end{array}\]
\end{definition}
$R_{\mathcal{K}}$ derives consequences from the rules in a way similar to the classical $T_P$ operator of definite logic programs, while
$D_{\mathcal{K}}$ obtains knowledge from the ontology $\mathcal{O}$, 
both from non-DL-atoms and the equalities occurring in
$\mathcal{O}$, where the $\approx$ operator defines a congruence relation between individuals.

The operator $T_{\mathcal{K}}$ is shown to be monotonic in \cite{KAH:ECAI08} so, by the Knaster-Tarski theorem,
 it has a unique least fixed point, denoted $\lfp(T_{\mathcal{K}})$, which is reached after a finite number of iteration steps.

The computation of the well-founded models follows the alternating fixed point construction \cite{AvG89} of the well-founded semantics for logic programs.  
This approach requires turning a hybrid MKNF knowledge base into a positive one to make $T_{\mathcal{K}}$ applicable.
\begin{definition}\label{d:MKNFtransformK}
Let $\mathcal{K}_{G}=(\mathcal{O},\mathcal{P}_{G})$ be a ground DL-safe hybrid MKNF knowledge base and let $S\subseteq{\sf KA}(\mathcal{K}_{G})$.
The \emph{MKNF transform} $\mathcal{K}_{G}/S = (\mathcal{O},\mathcal{P}_{G}/S)$ is obtained by $\mathcal{P}_{G}/S$ containing all rules 
$\K H\leftarrow \K A_{1},\ldots, \K A_{n}$ for which there exists a rule $$\K H\leftarrow \K A_{1}, \ldots, \K A_{n}, \nf B_{1},\ldots, \nf B_{m}$$ in $\mathcal{P}_{G}$ with $\K B_{j}\not\in S$ for all $1\leq j\leq m$.
\end{definition}
The above transformation  resembles  
that used for answer-sets \cite{answer-sets} of logic programs and the following two operators are defined. 
\begin{definition}\label{d:GammaK}
Let $\mathcal{K}=(\mathcal{O},\mathcal{P})$ be a
DL-safe hybrid MKNF knowledge base and $S\subseteq
\sf{KA}(\mathcal{K}^{*})$. We define:
$$\Gamma_{\mathcal{K}}(S) = \lfp(T_{\mathcal{K}^{+}_{G}/S})\mbox{,\ \  and\ \  }\Gamma'_{\mathcal{K}}(S) = \lfp(T_{\mathcal{K}^{*}_{G}/S})$$
\end{definition}

Both operators are shown to be antitonic \cite{KAH:ECAI08}, hence their composition is monotonic and form the basis for defining the well-founded MKNF model.
Here we present its alternating computation.

\begin{center}
\begin{tabular}{r @{=} p{2.5cm} r @{=} l}
$\mathbf{T}_{0}$ $ $ & $ $ $\emptyset$ & $\mathbf{TU}_{0}$ $ $ & $ $ $\sf{KA}(\mathcal{K}^{*})$\\
$\mathbf{T}_{n+1}$ $ $ & $ $ $\Gamma_{\mathcal{K}}(\mathbf{TU}_{n})$ & $\mathbf{TU}_{n+1}$ $ $ & $ $ $\Gamma'_{\mathcal{K}}(\mathbf{T}_{n})$ \\
$\mathbf{T}_{\omega}$ $ $ & $ $ $\bigcup\mathbf{T}_{n}$ & $\mathbf{TU}_{\omega}$ $ $ & $ $ $\bigcap\mathbf{TU}_{n}$
\end{tabular}
\end{center}
Note that by finiteness of the ground knowledge base
(Definition~\ref{d:MKNFKB}) the iteration stops before reaching
$\omega$.
It was shown in \cite{KAH:ECAI08} that the sequences are monotonically increasing and decreasing respectively, and that $\mathbf{T}_{\omega}$ and $\mathbf{TU}_{\omega}$ form the well-founded model in the following sense:

\begin{definition}\label{d:MKNFWFS}
Let $\mathcal{K}=(\mathcal{O},\mathcal{P})$ be a DL-safe hybrid MKNF knowledge base and let $\mathbf{T}_{\mathcal{K}}, \mathbf{TU}_{\mathcal{K}} \subseteq \sf{KA}(\mathcal{K})$, with $\mathbf{T}_{\mathcal{K}}$ being $\mathbf{T}_{\omega}$  and $\mathbf{TU}_{\mathcal{K}}$ being $\mathbf{TU}_{\omega}$ both restricted to the modal atoms occurring in $\sf{KA}(\mathcal{K})$.
Then 
$$M_{WF} = \{\K A\mid A\in \mathbf{T}_{\mathcal{K}}\}\cup \{\K \pi(\mathcal{O})\} \cup \{\nf A \mid A\in \sf{KA}(\mathcal{K}) \setminus\mathbf{TU}_{\mathcal{K}}\}$$ 
is the \emph{well-founded MKNF model} of $\mathcal{K}$, where $\pi(\mathcal{O})$ denotes the first-order logic formula equivalent to the ontology  $\mathcal{O}$ (for details on the translation of  $\mathcal{O}$ into first-order logic see \cite{MotikR07}) .
\end{definition}
All modal $\mathbf{K}$-atoms in $M_{WF}$ are true, all modal $\mathbf{not}$-atoms are false and all other modal atoms from $\sf{KA}(\mathcal{K})$ are undefined. 

As shown in \cite{KAH:ECAI08}, the well-founded model is sound with respect to the original semantics of \cite{MotikR07}, i.e. all atoms true (resp. false) in the well-founded model are also true (resp. false) in the model of \cite{MotikR07}. In fact, the relation between the semantics of  \cite{KAH:ECAI08} and \cite{MotikR07}, is tantamount to that of the well-founded semantics and the answer-sets semantics of logic programs. Moreover, this definition is in fact a generalization of the original definition of the well-founded semantics of normal logic programs, in the sense that if the ontology is empty then this definition exactly yields the well-founded models according to \cite{GRS91}.
For more properties, as well as motivation and intuitions on \MKNF{}, the reader is referred to \cite{KAH:ECAI08}.

\section{XSB Prolog and the Coherent Description Framework}\label{xsb}
Our implementation of \MKNF{} is based on XSB Prolog ({\tt
  xsb.sourceforge.net}) for two reasons.  First, XSB's tabling engine
evaluates rules according to WFS, and ensures rule termination for
programs and goals with the {\em bounded term-size
  property}~\footnote{Intuitively, a program $P$ and goal $G$ have the
  bounded term-size property if there is a finite number $n$ such that
  all subgoals and answers created in the evaluation of the goal $G$
  to $P$ have a size less than $n$.}.  Second, the implementation uses
the prover from XSB's ontology management system, the Coherent
Description Framework (CDF)~\cite{cdf}.

CDF has been used in numerous commercial projects, and was originally
developed as a proprietary tool by the company XSB, Inc~\footnote{Most
  of CDF is open-source, including all features used in this paper.
  CDF is distributed as a package in XSB's standard release, and full
  details can be found in its accompanying manual.}.  Since 2003, CDF
has been used to support extraction of information about aircraft
parts from free-text data fields, about medical supplies and
electronic parts from web-sites and electronic catalogs, and about the
specifics of mechanical parts from scanned technical drawings.  Also,
CDF is used to maintain screen models for graphical user interfaces
that are driven by XSB and its graphics package, XJ {\tt
  (www.xsb.com/xj.aspx)}.  We discuss features of CDF that are
relevant to the implementation described in
Section~\ref{sec:description}.

\paragraph{Type-0 and Type-1 Ontologies}
All classes in CDF are represented by terms of the form {\em
  cid(Identifier, Namespace)}, instances by terms of the form {\em
  oid(Identifier, Namespace)}, and relations by terms of the form {\em
  rid(Identifier, Namespace)}, where {\em Identifier} and {\em
  Namespace} can themselves be any ground Prolog term.  

Commercial use has driven CDF to support efficient query answering
from Prolog for very large knowledge bases.  A key to this is that
ontologies in CDF can have a restricted, tractable form.  {\em Type-0}
ontologies do not allow representation of negation or disjunction
within the ontology itself, and implicitly use the closed-world
assumption.  As such, Type-0 ontologies resemble a frame-based
representation more than a description logic, and do not add any
complexity to query evaluation beyond that of WFS.
Support for query answering motivates the representation of Type-0
ontologies.  The predicate $isa/2$ is used to state inclusion: whether
the inclusion is a subclass, element of, or subrelation depends on the
type of the term (not all combinations of types of terms are allowed
in a CDF program).  Relational atoms in CDF have the form:
\begin{itemize}
\item $hasAttr(Term_1,Rel_1,Term_2)$ which has the meaning $Term_1
\sqsubseteq \exists Rel_1.Term_2$;
\item $allAttr(Term_1,Rel_1,Term_2)$ with
the meaning $Term_1 \sqsubseteq \forall Rel_1.Term_2$;
\item along with
other forms that designate cardinality constraints on relations.
\end{itemize}
Figure~\ref{fig:type0} presents some DL Axioms and their Type-0
counterparts, where namespace information has been omitted for
readability.
\begin{figure} 
\longline
\begin{tabbing}
fooooooooooooooooooooooooo\=foo\=\kill
$Man \sqsubseteq Person \sqcap Male$ \\
    \> {\tt isa(cid(man),cid(person))} \\
    \> {\tt isa(cid(man),cid(male))} \\
\\
$Husband \sqsubseteq Man \sqcap \exists Spouse.Person$ \\
    \> {\tt isa(cid(husband),cid(man)).} \\
    \> {\tt hasAttr(cid(husband),rid(spouse),cid(person)} \\
\\
$adam : Husband$  \\
    \> {\tt isa(oid(adam),cid(husband)}
\end{tabbing}
\longline
\caption{Some DL Axioms and their Type-0 Counterparts}
\label{fig:type0}
\end{figure}
The fact that Type-0 ontologies cannot express negation is crucial to
their ability to be directly queried.  In other words, {\tt isa/2},
{\tt hasAttr/3} and other atoms can be called as Prolog goals, with
any instantiation pattern for the call.  Query answering for Type-0
goals checks inheritance hierarchies and does not rely on unification
alone. 
Type-0 ontologies use tabling to implement inheritance and use tabled
negation so that only the most specific attribute types for a {\tt
  hasAttr/3} or other query are returned to a user.

Besides the constructs of Type-0 ontologies, Type-1 ontologies further allow atoms of the form
$$necessCond(Term_1,CE)$$ where $CE$ can be any $\mathcal{ALCQ}$ class
expression over CDF terms. For instance, the axiom
\[
   Woman \sqsubseteq Person \sqcap \neg Man
\]
would be represented by the atom
\begin{verbatim}
  necessCond(cid(woman),(cid(person),neg(cid(man))))
\end{verbatim}
where the comma represents conjunction, as in Prolog.  Because they
use open-world negation, atoms for Type-1 ontologies cannot be
directly queried; rather they are queried through goals such as
$allModelsEntails(Term,ClassExpr)$, succeeding if $Term \sqsubseteq
ClassExpr$ is provable in the current state of the ontology.  Type-1
ontologies deduce entailment using a tableau prover written in Prolog.

\paragraph{System Features of CDF}
Regardless of the type of the ontology, atoms such as $isa/2$,
$hasAttr/2$, etc. can be defined extensionaly via Prolog atoms, or
intensionaly via Prolog rules.  For instance, evaluation of the goal 
\begin{verbatim}
  hasAttr(Class1,Rel,Class2)
\end{verbatim}
would directly check extensional Prolog facts through a subgoal 
\begin{verbatim}
  hasAttr_ext(Class1,Rel,Class2)
\end{verbatim}
and would also check intensional rules through a subgoal 
\begin{verbatim}
  hasAttr_int(Class1,Rel,Class2)
\end{verbatim}
Intensional definitions are used so that atoms can be lazily defined
by querying a database or analyzing a graphical model: their semantics
is outside that of CDF.  In fact, using combinations of rules and
facts, Type-0 ontologies are commonly used comprising tens of
thousands of classes and relations, and tens of millions of statments
about individuals.  At the same time, intensional definitions in a
Type-1 ontology provide a basis for the tableau prover to call rules,
as is required to support the interdependencies of \MKNF{}, and will
be further discussed in Section~\ref{sec:components}.

Despite their restrictions, the vast majority of knowledge used by
XSB, Inc. is maintained in Type-0 ontologies.  Although Type-0
ontologies have supported numerous commercial projects, their
limitations of course preclude the full use of information in
ontologies.  Support of a uniform querying mechanism for individualss
in \MKNF{} as described below is intended as a means to allow
commercial projects to use a more powerful form of knowledge
representation.

\subsubsection*{Related Work}

CDF was originally developed in 2002-2003, and Type-0 ontologies were
envisioned as a means to represent object-oriented knowledge in
Prolog.  Unlike Flora-2~\cite{flora2} CDF was intended to be a Prolog
library, and its inheritance was designed to be entirely monotonic for
compatibility with description logics.  Type-1 ontologies were
originally evaluated through a translation into ASP~\cite{Swif04}:
this approach pre-dated that of KAON2~\cite{Mot06} and was abandoned
due to the difficulty of dynamically pruning search in ASP; afterwards
the current tableau approach was developed which attempts to examine
as small a portion of the ontology as possible when proving entailment
(cf. e.g. \cite{HoPS99} for a discussion of how search may be pruned
in tableau provers for ontologies).  CDF's approach may be
distinguished from \cite{LuSK08}, which also combines ontological
deduction with Prolog.  Type-0 ontologies rely on WFS reasoning and so
achieve good scalability under a weak semantics; theorem proving for
Type-1 ontologies is used only when needed; \cite{LuSK08} takes a more
uniform approach to deduction which relies on WAM-level extensions for
efficiency; to our knowledge this approach is research-oriented, and
has not been used commercially.

\section{Goal-Driven MKNF Implementation}~\label{sec:description}

In Section~\ref{s:prel} we presented a bottom-up computation that constructs
the complete well-founded model for a given hybrid knowledge base. 
However, in practical cases, especially when considering the context of the
Semantic Web, this is not what is intended. In fact, it would make little sense 
to compute the whole model of anything that is related to the \emph{World Wide} Web.
Instead, one would like to query the knowledge base for a given predicate
(or propositional atom) and determine its truth value.
As an illustration, recall Example~\ref{motivEx} where we wanted to
know if a given shipment should be inspected or not when it arrived,
or the case study of \cite{PatelEtAl07} where one may want to know
whether a given patient matches the criteria for a given trial.
Deriving all the consequences of a knowledge base to answer a query
about a given shipment or patient would be impractical.

In this section we describe the algorithms and the design of
\oursystem{}, a goal-driven implementation for Hybrid MKNF Knowledge
Bases under the Well-Founded Semantics that minimizes the computation
to the set of individuals that are relevant to a query.  \oursystem{}
makes use of XSB's tabled SLG Resolution \cite{CheW96} for the evaluation of
a query, together with tableaux mechanisms supported by CDF theorem
prover to check entailment on the ontology. \oursystem{} is tuned for Type-1 ontologies, and thereby is also compatible with Type-0 ontologies. 
For the description of the solution, we assume that the reader has a general knowledge of tabled logic programs (cf. e.g. \cite{SwiW12}).

\subsection{A Query-Driven Iterative Fixed Point}
At an intuitive level, a query to \oursystem{} is evaluated in a
relevant (top-down like) manner through SLG resolution,
until the selected goal is a literal $l$ formed over a
DL-atom.  At that point, in addition to further resolution, the
ontology also uses tableau mechanisms to derive $l$.  However, as a
tableau proof of $l$ may require propositions (literals) inferred by
other rules, considerable care must be taken to integrate the tableau
proving with rule-based query evaluation.

In its essence, a tableau algorithm decides the entailment of a
formula $\varphi$ w.r.t. an ontology $\mathcal{O}$ by trying to
construct a common \emph{model} for $\neg \varphi$ and $\mathcal{O}$,
sometimes called a \emph{completion graph}
(cf. e.g.~\cite{ShmS91}). If such a model can not be constructed,
$\mathcal{O} \models \varphi$; otherwise $\mathcal{O}$ does not entail
$\varphi$.  Similar to other description logic provers, the CDF
theorem prover attempts to traverse as little of an ontology as
possible when proving $\varphi$.  As a result, when the prover is
invoked on an atom $A$, the prover attempts to build a model for the
underlying individual(s) to which $A$ refers, and explores additional
individuals only as necessary.

For our purposes, given the particular interdependence between the
rules and the ontology in \MKNF{}, the prover must consider the
knowledge inferred by the rules in the program for the entailment
proof, as a DL-atom can be derived by rules, which in turn may rely on
other DL-atoms entailed by the ontology.  Thus, a query to a DL-atom
$p(o)$, iteratively computes a (sub-)model for $o$, deriving at each
iteration new information about the roles and classes of $o$, along
with information about other individuals related to $o$ either in the
ontology (via CDF's tableau algorithm) or in the rules (via SLG
procedures) until a fixed point is reached.

We start by illustrating the special case of positive knowledge bases
without default negation in the rules.  

\begin{example}
\label{ex1}
  Consider the following KB (with the program on the left and the
  ontology on the right\footnote{To simplify reading, for rules we omit the $\mathbf{K}$ before 
non-DL atoms.  In fact, in the implementation the ontology must be
written according to CDF syntax, and in the rules the modal operators
$\mathbf{K}$ and $\nf$ are replaced by (meta-)predicates $known/1$ and
$dlnot/1$, respectively (see Section~\ref{sec:components}).}) and the
  query $third(X)$:
\[\begin{array}{lll}
\mathbf{K} \ third(X) \leftarrow  p(X), \mathbf{K} \ second(X). \ \ \ \ \ \ \ \ \\ 
\mathbf{K} \ first(callback). &\ \ \ \ \ \ \ \ First \sqsubseteq Second\\
p(callback). \\
\end{array}\]

The query resolves against the rule for $third(X)$, leading to the
goals $p(X)$ and $\mathbf{K} \; second(X)$.  The predicate $p$,
although not a DL-atom, assures DL-safety, restricting the application
of the rules to known individuals. The call $p(X)$ returns true for $X
= callback$.  
Accordingly, the next subgoal is $\K second(callback)$
which depends on the DL-atom
$second(callback)$, corresponding in the ontology to the proposition
$Second$. At this point, the computation calls the CDF theorem prover which
starts to derive a model for all the properties of the individual
$callback$. Yet in this computation, the proposition $Second$ itself
depends on a predicate defined in the rules -- $First$.  It can thus be seen
that the evaluation of the query $third(callback)$ must be
done iteratively -- the (instantiated) goal $third(callback)$ should
suspend (using tabling) until $second(callback)$ is
resolved. Furthermore, $second(callback)$ needs first to prove
$first(callback)$ from the rules.  In general, goals to DL-atoms may
need to suspend in order to compute an iterated fixed point, after
which they may either succeed or fail.
\end{example}

To formalize the actions in Example~\ref{ex1} on the special case of
definite programs, we start by considering the computation for all
individuals (i.e. temporarily disregarding the relevance of
individuals, as discussed above).

\begin{definition}
\label{iter}
Let $\mathcal{K} = (\mathcal{O}, \mathcal{P})$ be a DL-safe hybrid
MKNF knowledge base, where $\mathcal{P}$ does not contain default
negation.  Let $\cI$ be a fixed set of individuals.  The function
{\em Tableau$(\mathcal{O})$} computes for a theory $\mathcal{O}$ the
entailments of $\mathcal{O}$ for $\mathcal{I}$, disregarding the rules
component.  The function $SLG(\mathcal{P})$ computes via tabling the
set of DL-atoms true in the minimal model of $\mathcal{P}$ for a set
of individuals, $\mathcal{I}$, disregarding the ontology component.
The model is obtained as the union of the least fixed point of the sequences:
\begin{align*}
&D_0 = Tableau(\mathcal{O}) \qquad & &R_0 = SLG(\mathcal{P}) \\
&D_1 = Tableau(\mathcal{O} \cup R_0 ) \qquad & &R_1 = SLG(\mathcal{P} \cup D_0) \\
&\ldots \qquad & &\ldots \\
&D_n = Tableau(\mathcal{O} \cup R_{n-1})  \qquad & &R_n = SLG(\mathcal{P} \cup D_{n-1})  
\end{align*}
\end{definition}

Definition~\ref{iter} resembles Definition~\ref{d:opRkDkTk} of the
operator $T_\mathcal{K}$ in Section~\ref{s:prel}.  In fact, the $R_i$
sequence is similar to the $R_{\mathcal{K}}(S)$ operator which
collects new conclusions from the rules, whereas the $D_i$ sequence is
similar to the $D_{\mathcal{K}}(S)$ operator which collects new
conclusions from the ontology. Rather than starting with the empty set
of conclusions from the rules, as is the case for $T_\mathcal{K}$ of
Definition~\ref{d:opRkDkTk}, here the $R_i$ sequence starts with all
conclusions that can be drawn from the program alone~\footnote{Since
  $T_\mathcal{K}$ is monotonic, it is clear that starting the
  iteration of this operator with all conclusions drawn from the rules
  alone would yield exactly the same (least) fixed point.}. Given
these minor differences, taking the function {\em
  Tableau$(\mathcal{O})$} as correct w.r.t. the consequence relation
of the description logic in use, and taking the function
$SLG(\mathcal{P})$ as correct w.r.t. the least model semantics of
definite logic programs, it is easy to see that if $\mathcal{I}$ is
the set of all individuals in the knowledge base then the union of the
$D_n$ and $R_n$ at the fixed point exactly coincides with the least
fixed point of $T_\mathcal{K}$. Furthermore, as long as the program
$P$ respects DL-safety, MKNF rules are lazily grounded with respect to
the set of individuals $\mathcal{I}$. In fact, given a DL-safe set of
MKNF rules, and a set of queries grounded with the individuals in
$\mathcal{I}$, the evaluation of the queries results in a complete
grounding of the rules, and so the fixed point is guaranteed in a
finite number of steps.


Definition~\ref{iter} captures certain aspects of how the rules and
the ontology use each other to derive new knowledge in
\oursystem{}, via an alternating computation between the rules and the
ontology.  However it does not capture cases in which the relevant set
of individuals changes (i.e. it does not deal with changes in the set
$\mathcal{I}$), or the presence of default negation in rule bodies.
With regard to relevant individuals, since it is possible both to define
n-ary predicates in rules and roles in the ontology, a query
may depend on several individuals.  Therefore, the fixed
point computation must take into account the entire set of individuals
that the query depends on. This is done by tabling information about
each individual in the set of individuals relevant to the query.  This
set may increase throughout the fixed point iteration as new
dependency relations between individuals (including equality) are
discovered.  The iteration stops when it is not possible to derive
anything else about these individuals, i.e., when both the set of
individuals and the classes and roles of those individuals have
reached a fixed point. The details of this iterative increase in the
set of considered individuals can be found in the algorithm of
Figure~\ref{topalgo}, which also addresses default negation and its
interplay with first-order negation.

The following example considers the presence of default negation in
rule bodies.

\begin{example}
\label{ex1not}
Consider the following knowledge base:
\[\begin{array}{ll}
\mathbf{K} \ fourth(X) \leftarrow p(X), \nf \ third(X). &
Fourth \sqsubseteq Fifth\\
\mathbf{K} \ third(X) \leftarrow p(X), \mathbf{K} \ second(X). \ \ \ \ \ \ \ \ \ \ \ \ \ \ \ \  & First \sqsubseteq Second \\
\mathbf{K} \ first(callback).\\
p(callback). 
\end{array}\]

In this example a predicate $fourth(X)$ is defined as the default
negation of $third(X)$. Since $fourth(X)$ is defined in the rules,
the negation is \emph{closed world}, that is, $fourth(X)$ should only
succeed if it is not possible to prove $third(X)$.  Consequently, if
we employed SLG resolution blindly, an iteration where the truth of
$second(callback)$ had not been made available to the rules from the
ontology might mistakenly fail the derivation of $third(callback)$ and
so succeed $fourth(callback)$.  Likewise, the rules may pass to the
ontology knowledge, that after some iterations, no longer applies ---
in this case if the ontology were told that $fourth(callback)$ was
true, it would mistakenly derive {\em Fifth}.
\end{example}

Example~\ref{ex1not} illustrates the need to treat default negation carefully,
as the truth of default literals requires re-evaluation when new knowledge is inferred. Recall
the manner in which the operators $\Gamma_{\mathcal{K}}$ and
$\Gamma'_{\mathcal{K}}$ of Definition~\ref{d:GammaK} address the problem of
closed-world negation. Roughly, one step in $\Gamma_{\mathcal{K}}$ (or  $\Gamma'_{\mathcal{K}}$) is
defined as the application of $T_\mathcal{K}$ until reaching a fixed
point. Applying $\Gamma'_{\mathcal{K}}$ followed by $\Gamma_{\mathcal{K}}$
is a monotonic operation and thus is guaranteed to have a least
fixed point. In each dual application of $\Gamma_{\mathcal{K}}$ and
$\Gamma'_{\mathcal{K}}$ two different models follow -- a monotonically
increasing model of true atoms (i.e. true predicates and propositions), and
a monotonically decreasing model of non-false atoms.

In a similar way, the implementation of \oursystem{} makes use of two
fixed points: an \emph{inner} fixed point where we apply
Definition~\ref{iter} corresponding to $T_\mathcal{K}$; and an
\emph{outer} fixed point for the evaluation of $\mathbf{not}$s, corresponding
to $\Gamma_\mathcal{K}$ (and $\Gamma'_{\mathcal{K}}$).  In the outer iteration, the evaluation of
closed-world negation is made by a reference to the previous model
obtained by $\Gamma_\mathcal{K}$.  Thus in \oursystem{}, $\mathbf{not}(A)$
succeeds if, in the previous \emph{outer} iteration, $A$ was not
proven.

\begin{example}\label{exnot}
To illustrate the need to apply two fixed points, consider the
knowledge base below and the query $c(X)$:
\[\begin{array}{ll}
\mathbf{K} \ c(X) \leftarrow p(X), \mathbf{K} \ a(X), \mathbf{not} \ b(X). \ \ \ \ \ \ \ \  \ \ \ \ \ \ \ \ & A \sqsubseteq B\\
\mathbf{K} \ a(object).\\
p(object).
\end{array}\]

When evaluating the query $c(X)$, $X$ is first bound to $object$ by
$p$, and then the nested iteration begins.  The inner iteration
follows the steps of Definition~\ref{iter}, and since these operators
are defined only for definite rules, each negative body literal in a
rule such as $p(X)$ is evaluated according to its value in the
previous {\em outer} fixed point, or is simply evaluated as {\em true}
in the first outer iteration.  (As we will see, this is done lazily by
\oursystem{}).  The first stage of the inner iteration computes $R_0 =
\{a(object), p(object),c(object)\}$ (Definition~\ref{iter}) via the
rules; and computes via the ontology $D_0 = \emptyset$, 
as $\mathcal{O} \not \models A$.
\comment{
When evaluating the query $c(X)$, $X$ is first bound to $object$ by
$p$, and then the iteration process of Definition~\ref{iter} begins.
Note that Definition~\ref{iter} refers only to definite programs.  To
treat a rule like that for $c(X)$ as positive, each negative body
literal is evaluated according to its value in the previous outer
fixed point, or is simply evaluated as true in the first outer
iteration.  As we will see, this is done lazily by \oursystem{}.
Accordingly, the rules infer $a(object)$, $p(object)$ and $c(object)$
for $R_0$.  However in the first inner iteration the set of
ontological entailments, $D_0$, is empty since $\mathcal{O} \not
\models A$.
}
In the second inner stage the rules achieve the same fixed point as in
the first, so $R_1 = R_0$, but the ontology derives $object:B$
in $D_1$. After sharing this knowledge, there is nothing else to infer
by either components, and we achieve the first inner fixed point with:
\[
T_1 = \{a(object),b(object),c(object),p(object)\}
\]
So now, the second outer iteration starts the computation of the inner
iteration again and, in this iteration, negative literals in the rules
are evaluated w.r.t. $T_1$.  As a consequence, $c(object)$ fails,
since $b(object) \in T_1$.  The fixed point of the second inner
iteration contains $p(object)$, $a(object)$ and $b(object)$, which is
in fact the correct model for $object$. Afterwards, a final outer
iteration is needed to to determine that an outer fixed point has in
fact been reached. Since $c(object)$ is in the model of the final
iteration, the query $c(X)$ succeeds for $X = object$.
\end{example}

The procedure to compute a lazily invoked iterative fixed point over a
DL-safe MKNF Hybrid Knowledge Base
is summarized in Figure~\ref{topalgo} using predicates that are
described in detail in Section~\ref{sec:components}.  In each inner
iteration, the tabled predicate $known/3$ is used to derive knowledge
from the rules component, while $allModelsEntails/3$ infers knowledge
from the ontology via tableau proofs.  Within rules evaluated by
$known/3$, the default negation of a DL-atom $A$ is obtained by the
predicate $dlnot(A)$, which succeeds if $A$ was not proven in the last
outer iteration.  The predicates $definedClass/2$ and $definedRole/3$
are used to obtain the relevant classes and roles defined for a given
individual.  We assume that these predicates are defined explicitly by
the compiler or programmer, but they can also be inferred via the
DL-safe restriction~\footnote{Because of DL-safety, every DL-rule must
  contain a positive literal that is {\em only} defined in the
  rules. Such a literal limits the evaluation of the rules to known
  individuals, so that \oursystem{} can infer the set of individuals
  that are applicable to a given rule.}.  Regardless of whether
inference is used, whenever a role is encountered for an individual, a
check is made to determine whether the related individual
$Individual_1$ is already in the list of individuals in the fixed
point, and $Individual_1$ is added if not.


\begin{algorithm}
\linesnumbered
\SetLine
\SetKwInput{Input}{Input}
\SetKwInput{Output}{Output}
\Input{A query $Query$ to a DL-Atom}
\Output{Value of the input query in \MKNF}
\SetKwFor{Foreach}{foreach}{do}{end}
\emph{addIndividuals($Query$,IndividualList)}; \\
\Foreach{ Individual \emph{\bf in} IndividualList}{
    \emph{OutIter}, \emph{InIter} = 0; \\ 
    $S = S_1 = \{ \}$; \\
    $P = P_1 =  \{ \}$; \\
    \Repeat{OutIter is even (a TU evaluation) and $P$ = $P_1$}{
    $P = P_1$; \\
    \Repeat{$S = S_1$}{
    $S = S_1$; \\
    \ForEach{ Class \emph{\bf in} definedClass(Individual,Class) }{
        \emph{Term} = \emph{Class(Individual)}; \\
        $S_1 = S_1 \cup$ \emph{known(Term, OutIter, InIter)}; \\
        $S_1 = S_1 \cup$ \emph{allModelsEntails(Term, OutIter, InIter)}; \\
        $S_1 = S_1 \cup$ \emph{allModelsEntails(neg(Term), OutIter, InIter)}; \\
        }
    \ForEach{ Role \emph{\bf in} definedRole(Individual,Individual$_1$,Role) }{
        \emph{Term} = \emph{Role(Individual,Individual$_1$)}; \\
        add \emph{Individual$_1$} to \emph{IndividualList} if necessary \\
        $S_1 = S_1 \cup$ \emph{known(Term, OutIter, InIter)}; \\
        $S_1 = S_1 \cup$ \emph{allModelsEntails(Term, OutIter, InIter)}; \\
        $S_1 = S_1 \cup$ \emph{allModelsEntails(neg(Term), OutIter, InIter)}; \\
        }
    \emph{InIter}++; \\ 
    }
    $P = S$; \\
    \emph{OutIter}++;
   }   
}
\eIf{known($Query$,Outer$_{Final}$-1,Inner\ $_{Final}$)}{\textbf{return} true}
    {
    \eIf{known($Query$,Outer$_{Final}$,Inner'$_{Final}$)}{\textbf{return} undefined}
    {\textbf{return} false}
    }

\caption{The Top-Level Algorithm:{\em ComputeFixedPoint(Query)}}
\label{topalgo}
\end{algorithm}

In order to compute \MKNF{} the algorithm shown in
Figure~\ref{topalgo} must create two different sets
(cf. Definition~\ref{d:MKNFWFS}): a credulous set, $TU$, containing
the atoms that are true or undefined corresponding to application of
the operator $\Gamma'_{\cal K}$ ; and a skeptical set, $T$, of the
atoms that are true corresponding to $\Gamma_{\cal K}$
(Definition~\ref{d:GammaK}).
\comment{
Finally, after computing the sets and
achieving the fixed point, the algorithm returns the evaluation of
$known(Query,Final_{Outer}-1)$, where $Final_{Outer}$ represents the iteration
where the outer fixed point was accomplished. Since the first outer
set obtained corresponds to the first iteration in the $TU$ set, this
outer fixed point will be obtained in a $TU$ iteration. Thus to check
if $Query$ is true, we need to check if it is contained in the set
inferred in $Final_{Outer}-1$.  If this is not the case, $Query$ is
evaluated as undefined if it is derived in $Final_{Outer}$, and as false
otherwise.
}
%
The iterations in Figure~\ref{topalgo} capture the construction of
these sets in the following manner.
The first iteration of $OutIter$ in $known(Term,OutIter,InIter)$
(where $OutIter = 0$) corresponds to the first step of computation of
the set $TU$, whereas the second iteration corresponds to the second
step of computation of the set $T$. As a consequence, iterations where
$OutIter$ is indexed with an even number are monotonically decreasing
$TU$ iterations, while iterations indexed with an odd number are
monotonically increasing $T$ iterations.  By making use of this
property, the algorithm of Figure~\ref{topalgo} ensures that the fixed
point will only be achieved in $TU$ iterations. This way, $Query$ is
true if $known(Query,Outer_{Final}-1,Inner_{Final})$ holds. If this is
not the case, then $Query$ is undefined if
$known(Query,Outer_{Final},Inner_{Final})$ holds, and $Query$ is false
otherwise.

\subsection{Implementing  \MKNF{} Components}~\label{sec:components}
%

We now describe the various predicates in the algorithm of
Figure~\ref{topalgo}, including the manner in which the rule and
ontology components exchange knowledge, and how the fixed point is
checked.

\subsubsection{Rules Component}
As mentioned, rules are transformed to use $known/1$ corresponding to
$\mathbf{K}$ and $dlnot/1$ corresponding to $\mathbf{not}$.
As shown in Figure~\ref{fig:known}, given the goal $known(A)$ with $A
= p(O)$, the code first calls $computeFixedPoint(p(O))$ to perform the
fixed point computation for the object instance $O$.  As in
Figure~\ref{topalgo}, $computeFixedPoint(p(O))$ calls the lower-level
$known/3$ and $dlnot/3$ to determine the truth of literals during the
fixed point computation.  Once the fixed point has been reached,
$known/1$ uses {\em get\_object\_iter(p(O),Outer,Inner)} to obtain the
final iteration indices for $O$ from a global store, and calls
$known/3$ again to determine the final truth value of $p(O)$.  Note
that $known/3$ is always called with the iteration indices (arguments
2 and 3) bound, so that they are always contained in the table
entries.  Thus, the post-fixed point call to $known/3$ simply checks
the table, and is not computationally expensive.

Within a given iteration, $p(O)$ may be known in one of two ways.
Either it can be directly derived from the rules; or
$O \in P$ (i.e. {\em o:P}) may have been entailed by the ontology in
the previous inner iteration step, as determined by the call
$allModelsEntails(p(O),OutIter, PrevIter)$.
In either case, care must be taken so that that if $\neg A$ holds,
then $\nf A$ holds as well. In the formalism of
Definition~\ref{d:kb+*} this is guaranteed in two steps.  First, an
axiom $\neg P \sqsubseteq \textit{NP}$ is added for each DL-atom that
occurs in the head of a rule; in addition, the literal $\nf NP$ is
added the body of each rule with head $P$: this
rewrite is used by the $\Gamma'_{\cal K}$ operator to produce the $TU$
set.  Accordingly in Figure~\ref{fig:known}, when we try to derive
$known(A,OutIter,InIter)$ and
$OutIter$ is even (i.e. corresponding to a $TU$ step via
$\Gamma_\mathcal{K}'$ in Definition~\ref{d:MKNFWFS} )
we check whether the ontology derived $\neg A$ in the previous inner
iteration by the call $no\_prev\_neg(A,OutIter,PrevIter)$.  If $\neg
A$ was derived, then $no\_prev\_neg/3$ fails via the call to $tnot/1$,
(which is XSB's operator for tabled negation), and the top-level goal
also fails.

\begin{figure}
\longline
\begin{lstlisting}
known(A):- 
     computeFixedPoint(A), 
     get_object_iter(A,OutIter,InIter),
     known(A,OutIter,InIter).

:- table known/3.
known(A,OutIter,InIter):- 
  PrevIter is InIter - 1,
  (   call(A),                       
    ; 
      InIter > 0,
      allModelsEntails(A,OutIter,PrevIter) 
  ),      
  ( OutIter mod 2 =:= 1 -> 
        true
      ;
        no_prev_neg(A,OutIter, PrevIter) 
  ).

/* Enforce coherence of default negation with first-order negation */	
no_prev_neg(_A,_OutIter, PrevIter) :- 
  PrevIter < 0,!.
no_prev_neg(A,OutIter, PrevIter) :- 
  tnot(allModelsEntails(neg(A),OutIter,PrevIter)). 	
\end{lstlisting}
\longline
\caption{Prolog Implementation of $\mathbf{K}$ for Class Properties}
\label{fig:known}
\end{figure}

\begin{figure}
\longline
\begin{lstlisting}
dlnot(A):- 
     computeFixedPoint(A),
     get_object_iter(A,OutIter,_InIter), 
     dlnot(A,OutIter).

/* In first iteration, ensure that TU = KA(K*) */
dlnot(_A,0):- !. 
/* In subsequent iterations, check previous outer iteration */
dlnot(A,OutIter):- 
        PrevIter is OutIter - 1, 
        get_final_iter(A,PrevIter,FinIter),
        tnot(known(A,PrevIter,FinIter)).
\end{lstlisting}
\longline
\caption{Prolog Implementation of $\mathbf{not}$ for Class Properties}
 \label{fig:dlnot}
 \end{figure}

On the other hand, the predicate $dlnot(A)$ which uses closed world
assumption, succeeds if $A$ fails (Figure~\ref{fig:dlnot}).  As
discussed in Example~\ref{exnot}, the evaluation of $dlnot/2$ must
take into account the result of the previous {\em outer} iteration.
Accordingly, in Figure~\ref{fig:dlnot} the call $dlnot(A)$ with $A =
p(O)$ gets the current outer iteration for $O$, and immediately calls
$dlnot/2$.  If the outer iteration index is greater than 1, the second
clause of $dlnot/2$ simply finds the index of the (inner) fixed point
of the previous outer iteration, and determines whether $A$ was true
in that fixed point.  Since the call to $known/3$ in $tnot/1$ is
tabled, {\em dlnot/[1,2]} do not need to be tabled themselves.
As described before, outer iterations alternately represent iterations
of $T$ and $TU$ sets of Definition~\ref{d:MKNFWFS}, where $T$ sets are
monotonically increasing while $TU$ sets are monotonically
decreasing. To assure that the first $TU$ set is the largest set ({\sf
  KA($\cK^*$)} following Definition~\ref{d:GammaK}), we compel all
calls to $dlnot/1$ to succeed in the first outer iteration, as
represented by the first clause of $dlnot/2$.

\subsubsection{Ontology Component}
The tabled predicate {\em allModelsEntails/3} provides the interface
to CDF's tableau theorem prover (Figure~\ref{fig:ame}).  It is called
with an atom or its classical negation, and with the its iteration
indices bound.  Although the iterations are not used in the body of
{\em allModelsEntails/3}, representing the iteration information in
the head ensures its availability in table entries for use by {\em
  known/3}.
{\em allModelsEntails/3} first converts the atomic form of a
proposition to one used by CDF.  That is, it translates a 1-ary
DL-atom representing an individual's class membership to the CDF
predicate $isa/2$, and a 2-ary DL-atom representing an individual's
role to the CDF predicate $hasAttr/3$ (see Section~\ref{xsb}).  In
addition, if $Atom$ is a 2-ary role, the target individual may be
added to the fixed point set of individuals.  As is usual with tableau
provers, entailment of a formula $\varphi$ by an ontology $\cO$ is
shown if the
\begin{figure}
\longline
\begin{lstlisting}
    :- table allModelsEntails/3.
    allModelsEntails(neg(Atom),_OutIter,_InIter):- !,
            /* transform Atom to CDF object identifier and class expression*/
            /* add individuals to current fixed point list */ 
            (rec_allModelsEntails(Id,CE) -> fail ; true).
    allModelsEntails(Atom,_OutIter,_InIter):-
            /* transform Atom to CDF object identifier and class expression*/
            /* add individuals to current fixed point list */
            (rec_allModelsEntails(Id,neg(CE)) -> fail ; true).
\end{lstlisting}
\longline
\caption{Prolog Pseudo-code for $allmodelsEntails/3$}
\label{fig:ame}
\end{figure}
classical negation of $\varphi$ is inconsistent with $\cO$.  Thus, {\em
  rec\_alModelsEntails/2} immediately fails if the classical negation
of $\varphi$ is consistent with $\cO$ in the present iteration; otherwise,
$\varphi$  is entailed.

The tableau prover, called by $rec\_allModelsEntails/2$,
obtains all information inferred by the rules during the previous
inner iteration, in accordance with Definition~\ref{iter}.  This is
addressed via the CDF intensional rules.  As discussed in
Section~\ref{xsb}, the architecture of a CDF instance can be divided
into two parts -- extensional facts and intensional rules. Extensional
facts define CDF classes and roles as simple Prolog facts; intensional
rules allow classes and roles to be defined by Prolog rules that are
outside of the \MKNF{} semantics.  In our case, the intensional rules
support a programming trick to check rule results from a previous
iteration.  As shown in Figure~\ref{fig:callback} they convert the CDF
form of an ontology axiom into a 1-ary or 2-ary predicate, and then
check the $known/3$ table for a previous iteration using the predicate
{\em lastKnown/1} (not shown).  If roles or classes are uninstantiated
in the call from the tableau prover, all defined roles and classes for
the individual are instantiated using {\em definedClass/3} or {\em
  definedRole/4} before calling {\em lastKnown/1}.

\begin{figure} 
\longline
\begin{lstlisting}
isa_int(oid(Obj,NS),cid(Class,NS1)):- 
    ground(Obj),ground(Class),!,
    Call =.. [Class,Obj],          /* Call = Class(Obj) */
    lastKnown(Call).
/* Find all possible classes for Obj if called with superclass argument uninstantiated */
isa_int(oid(Obj,NS),cid(Class,NS)):- 
    ground(Obj),var(Class),!,
    definedClass(Call,Class,Obj), 
    lastKnown(Call).

hasAttr_int(oid(Obj1,NS),rid(Role,NS1),oid(Obj2,NS2)):-
    ground(Obj1), ground(Obj2), ground(Role),!,
    Call =.. [Role,Obj1,Obj2],     /* Call = Role(Obj1,Obj2) */
    last_known(Call).
/* Find all possible rules for Obj if called with role argument uninstantiated */
hasAttr_int(oid(Obj1,NS),rid(Role,NS1),oid(Obj2,NS2)):-
    ground(Obj1), ground(Obj2), var(Role),!,
    definedRole(Call,Role,Obj1,Obj2),
    last_known(Call).
\end{lstlisting}
\longline
\caption{Callbacks from the ontology component to the rules component}
\label{fig:callback}
\end{figure}

\subsubsection{Usage}
An MKNF Hybrid Knowledge base is defined over a XSB-Prolog knowledge base together with an ontology specified over CDF. In \oursystem{} such a knowledge base is written into two files as follows: 
\begin{itemize}
\item $rules.P$ -- containing the set of MKNF rules and facts. A rule is defined as standard Prolog rules as follows:
{\small
   \begin{center}
      \texttt{  Head :- $A_1$, ... $A_k$,known($B_1$),$\ldots$,known($B_n$),dlnot($C_1$),$\ldots$,dlnot($C_m$). }
   \end{center} 
} where $k, n,m \geq 0$, and the $A_i$s are all non-DL predicates
(i.e. predicates that are not defined in the ontology), and the $B_i$s
and $C_i$s are predicates that can be both defined in the rules and in
the ontology. If $k =n=m=0$ then the rule is a fact, and it is written
as usual in Prolog, omitting the `:-' operator.  Note that the
transformation to include the negation {\em N\_Head} in the body of a
rule for {\em Head} as specified in Definition~\ref{d:kb+*} is not
needed: such a check is done by the call to {\tt no\_prev\_neg/3} in
{\tt known/3}.

To guarantee correctness, each rule must respect DL-safety.  However,
in the current implementation it is the programmer's responsibility to
check for this condition.
%
The current implementation also does not check that the $A_i$
predicates (i.e. the ones not under $known/1$ or $dlnot/1$) are not
defined in the ontology. If a programmer opts to not precede the
predicate by $known/1$ or $dlnot/1$, any definition for the
predicate in the ontology is simply ignored.

\item $cdf\_extensional.P$ -- comprising ordinary ontology facts and concepts defined over the CDF syntax. 

\item $cdf\_intensional.P$ -- containing predicates allowing the
  ontology to access information in the rules as in
  Figure~\ref{fig:callback}.  In addition, the file may contain other
  intensional rules to lazily access information from a database, off
  of the semanitc web, or from other sources external to \oursystem{}.
\end{itemize}

\comment{
Besides these two files defining MKNF hybrid knowledge bases,
currently \oursystem{} also allows for the definition of so called
\emph{CDF intensional rules}, to be written in a file
$cdf\_intensional.P$. Intensional rules are just usual Prolog rules,
where the head is a CDF formula. Moreover, $Body$ can also call back
directly any (n-ary) predicate from the ontology via the use of
$known/1$ and its negation with $dlnot/1$.
}

\begin{example}
The knowledge base of Example~\ref{motivEx}  can be easily coded in \oursystem{} as:
{\small 
\begin{lstlisting}
%rules
inspect(X) :- hasShipment(X,C), dlnot(safeCountry(C)).

%cdf_extensional
isa_ext(cid(scandinavianCtr,ont),cid(safeCountry,ont)).
isa_ext(cid(scandinavianCtr,ont),cid(EuropeanCtr,ont)).
isa_ext(oid(norway,ont),cid(scandinavianCtr,ont)).
necesscond_ext(cid(DiplomaticShipt,ont),neg(cid(inspect,ont))).
\end{lstlisting}
}
\noindent
Note that the ontological portion is Type-1, due to the use of {\tt
  necessCond/2}.
\end{example}

\subsubsection{Discussion}
As described, \oursystem{} implements query answering to hybrid MKNF
knowledge bases, and tries to reduce the amount of relevance required
in the fixed point operation.  Relevance is a critical concept for
query answering in practical systems, however a poorly designed
ontology or rules component can work against one another if numerous
individuals depend on one another through DL roles.  In such a case
the relevance properties of our approach will be less powerful;
however in such a case, a simple query to an ontology about an
individual will be inefficient in itself.  The approach of
\oursystem{} cannot solve such problems; but it can make query
answering as relevant as the underlying ontology allows.

We do not present here a formal proof of soundness and completeness
for our algorithm, since this would require the full presentation of
the formal derivation procedures on which both XSB-Prolog and CDF
implementations rely.  However, we have given an informal argument
along with our description by referring to complementarity between the
implementation and the bottom-up definition of \MKNF{}. In particular,
there is a close correspondence between the inner fixed points of our
computation represented in Definition~\ref{iter} and the $R_{\cal K}$,
$D_{\cal K}$ and $T_{\cal K}$ operators of
Definition~\ref{d:opRkDkTk}; a correspondance between the the actions
of {\tt known/3} in Figure~\ref{fig:known} and the transformation of
Definition~\ref{d:MKNFtransformK} to ensure the coherency between
classical and default negation; and also a correspondance between our
outer fixed points and the operators $\Gamma_{\cal K}$ /
$\Gamma'_{\cal K}$.  As a result, one can view our goal-driven
implementation as an optimization of the bottom-up approach where the
computation is limited to the set of \emph{relevant} objects, and
where the evaluation of positive predicates and the handling of
iterations is performed by the use of SLG resolution.

Further optimizations of the described approach are possible.  First
is to designate a set of atoms whose value is defined {\em only} in
the ontology: such atoms would require tableau proving, but could
avoid the fixed point check of $computeFixedPoint/1$.  Within
$computeFixedPoint/1$ another optimization would be to maintain
dependencies among individuals.  Intuitively, if individual $I_1$
depended on individual $I_2$ but not the reverse, a fixed point for
$I_2$ could be determined before that of $I_1$.  However, these
optimizations are fairly straightforward elaborations of \oursystem{}
as presented.

\section{Conclusions}

In this paper we have described the implementation of a query-driven
system, \oursystem{}, for hybrid knowledge bases combining both
(non-monotonic) rules and a (monotonic) ontology. The system answers
queries according to \MKNF{} \cite{KAH:ECAI08} and, as such, is also
sound w.r.t. the semantics defined in \cite{MotikR07} for
Hybrid MKNF knowledge bases. The definition of \MKNF{} is parametric
on a decidable description logic (in which the ontology is written),
and it is worth noting that, as shown in \cite{KAH:ECAI08}, the
complexity of reasoning in \MKNF{} is in the same class as that in the
decidable description logic; a complexity result that is extended to a
query-driven approach in~\cite{AlKS12}.  In particular, if the
description logic is tractable then reasoning in \MKNF{} is also
tractable. Our implementation fixes the description logic part to CDF
ontologies that, in its Type-1 version, supports $\mathcal{ALCQ}$
description logic. CDF Type-0 ontologies are simpler, and tractable
and, when using Type-0 ontologies only, our implementation exhibits a
polynomial complexity behavior. This fact derives from the usage of
tabling mechanisms, as defined in SLG resolution and implemented in
XSB Prolog\footnote{The proof of tractability of the implementation of \oursystem{} with CDF Type-0 ontologies is beyond the scope of this
paper.} In fact, one of the reasons that highly influenced the choice of CDF as
the parameter ontology logic in our query-driven implementation for Hybrid MKNF knowledge bases, was the very existence of an implementation of CDF relying on tabling, that could be coupled together with the tabling we needed for \MKNF{}. But the algorithms presented here do not rely on particularities of CDF, and we believe that, for other choices of parameter logics, implementations could be made in a way similar to the one described in this paper. Of course, such an implementation would require first an implementation in XSB-Prolog of a prover for the other description logic of choice, providing at least a predicate $allModelsEntails/3$ with the meaning as described above.

Though our choice for the implementation was the Well-Founded Semantics for Hybrid MKNF knowledge bases, \MKNF{}, \cite{KAH:ECAI08}, there were other formalisms concerned with combining ontologies with WFS rules~\cite{ELST04:RuleML,DM07}. The approach of \cite{ELST04:RuleML} combines ontologies and rules in a modular way, i.e. keeps both parts and their semantics separate, thus having similarities with \MKNF{}. The interface for this approach is done by the \texttt{dlv-hex} system \cite{dlvhex}. Though with identical data complexity to \MKNF{} for a tractable DL, it has a less strong integration, having limitations in the way the ontology can call back program atoms (see \cite{ELST04:RuleML}  for details).
Hybrid programs of \cite{DM07} are even more restrictive:
this formalism only allows the transfer of information from the ontology to the rules and not the other way around.
Moreover, the semantics of this approach differs from MKNF (both the one of  \cite{MotikR07} and \MKNF{}) and also\cite{ELST04:RuleML} in that if an ontology expresses $B_{1} \vee B_{2}$ then the semantics in \cite{DM07} derives $p$ from rules $p\leftarrow B_{1}$ and $p\leftarrow B_{2}$, $p$ while MKNF and \cite{ELST04:RuleML} do not.
For further comparisons of MKNF with other proposals, including those not based on WFS rules, see \cite{MotikR07,KAH:ECAI08}, and for a survey on other proposals for combining rules and ontologies see \cite{Hitzler-Parsia-OHbook}

\oursystem{} serves as a proof-of-concept for querying \MKNF{}
knowledge bases.  As discussed, XSB and tractable CDF ontologies have
been used extensively in commercial semantic web applications; the
creation of \oursystem{} is a step towards understanding whether and
how \MKNF{} can be used in such applications.  As XSB is
multi-threaded, \oursystem{} can be extended to a \MKNF{} server in a
fairly straightforward manner.  Since XSB supports CLP, further
experiments involve representing temporal or spatial information in a
hybrid of ontology, rules, and rule-based constraints.  In addition,
since the implementation of Flora-2~\cite{flora2} and
Silk~\cite{Gros09} are both based on XSB, \oursystem{} also forms a basis
for experimenting with \MKNF{} on these systems.

\bibliographystyle{acmtrans}

\bibliography{cdfrules}

\end{document}